\documentclass{article} 
\usepackage{iclr2025_conference,times}

\usepackage{amsmath,amsfonts,bm}

\def\eqref#1{equation~\ref{#1}}

\def\1{\bm{1}}

\DeclareMathAlphabet{\mathsfit}{\encodingdefault}{\sfdefault}{m}{sl}
\SetMathAlphabet{\mathsfit}{bold}{\encodingdefault}{\sfdefault}{bx}{n}

\usepackage{hyperref}
\usepackage{url}
\usepackage{graphicx}
\usepackage{wrapfig}
\usepackage{placeins}
\usepackage{xcolor}
\usepackage{booktabs}
\usepackage{multirow}
\usepackage{enumitem}

\definecolor{glaucous}{rgb}{0.38, 0.51, 0.71}
\definecolor{pastelmagenta}{rgb}{0.96, 0.6, 0.76}
\definecolor{orchid}{rgb}{0.85, 0.44, 0.84}

\hypersetup{
  colorlinks=true,
  citecolor=glaucous,
  linkcolor=pastelmagenta,
  urlcolor=orchid}

\usepackage{arydshln}
\makeatletter
\def\adl@drawiv#1#2#3{%
        \hskip.5\tabcolsep
        \xleaders#3{#2.5\@tempdimb #1{1}#2.5\@tempdimb}%
                #2\z@ plus1fil minus1fil\relax
        \hskip.5\tabcolsep}
\newcommand{\cdashlinelr}[1]{%
  \noalign{\vskip 1.3pt
           \global\let\@dashdrawstore\adl@draw
           \global\let\adl@draw\adl@drawiv}
  \cdashline{#1}[.4pt/2pt]
  \noalign{\global\let\adl@draw\@dashdrawstore
           \vskip 3pt}}
\makeatother

\title{\centering EuroLLM \\ Multilingual Language Models for Europe}

\author{
\vspace{0.3cm}
\bf
 Pedro Henrique Martins$^{1}$ \hspace{0.1cm}
 Patrick Fernandes$^{2,3}$ \hspace{0.1cm}
 João Alves$^{1}$ \hspace{0.1cm}
 Nuno M. Guerreiro$^{1,2,4}$ 
 \\
\bf
 Ricardo Rei$^{1}$ \hspace{0.1cm}
 Duarte M. Alves$^{2}$ \hspace{0.1cm}
 José Pombal$^{1,2}$ \hspace{0.1cm}
 Amin Farajian$^{1}$ \hspace{0.1cm}
 Manuel Faysse$^{4,5}$ 
 \\
\bf
 Mateusz Klimaszewski$^{6}$ \hspace{0.1cm}
 Pierre Colombo$^{4,7}$ \hspace{0.1cm}
 Barry Haddow$^{6,8}$ \hspace{0.1cm}
 José G. C. de Souza$^{1}$ 
 \\
\bf
 Alexandra Birch$^{6,8}$ \hspace{0.1cm}
 André F. T. Martins$^{1,2}$
\\
\vspace{0.3cm}
\normalfont
$^{1}$Unbabel \hspace{0.1cm} $^{2}$Instituto de Telecomunicações, Instituto Superior Técnico \\ 
$^{3}$Carnegie Mellon University \hspace{0.1cm} $^{4}$MICS, CentraleSupélec, Université Paris-Saclay\\
$^{5}$Illuin Technology \hspace{0.1cm} $^{6}$University of Edinburgh \hspace{0.1cm}  $^{7}$Equall \hspace{0.1cm} $^{8}$Aveni
}

\newcommand\blfootnote[1]{%
  \begingroup
  \renewcommand\thefootnote{}\footnote{#1}%
  \addtocounter{footnote}{-1}%
  \endgroup
}

\iclrfinalcopy 
\begin{document}

\maketitle

\begin{abstract}
The quality of open-weight LLMs has seen significant improvement, yet they remain predominantly focused on English. In this paper, we introduce the \emph{EuroLLM} project, aimed at developing a suite of open-weight multilingual LLMs capable of understanding and generating text in all official European Union languages, as well as several additional relevant languages. We outline the progress made to date, detailing our data collection and filtering process, the development of scaling laws, the creation of our multilingual tokenizer, and the data mix and modeling configurations. Additionally, we release our initial models: EuroLLM-1.7B and EuroLLM-1.7B-Instruct\footnote{The EuroLLM models are available \href{https://huggingface.co/collections/utter-project/eurollm-66b2bd5402f755e41c5d9c6d}{here}.} and report their performance on multilingual general benchmarks and machine translation.\blfootnote{Corresponding author: pedro.martins@unbabel.com.}
\end{abstract}

\section{Introduction}
Large language models (LLMs) are driving significant advancements in natural language processing and AI, as demonstrated by OpenAI’s GPT series and Anthropic's Claude. LLMs are pre-trained on vast amounts of unlabelled data to perform a self-supervised task (\textit{e.g.,} next word prediction or missing word prediction), enabling them to develop a deep understanding of language. These pre-trained LLMs can already perform various downstream tasks, often leveraging in-context learning techniques, but are typically fine-tuned to better follow natural language instructions, improve performance on specific tasks, and adhere to safety protocols.

However, the most advanced models are owned by large corporations with a piecemeal commitment to open science. Moreover, despite the growing availability of open-weight LLMs (\textit{e.g.,} LLaMA, Mistral, or Gemma \citep{touvron2023llama,jiang2023mistral,team2024gemma}), these are predominantly limited to English and a few high-resource languages, leaving out many European languages. To address this gap, we have started the EuroLLM project with the goal of creating a suite of LLMs capable of understanding and generating text in all European Union languages (Bulgarian, Croatian, Czech, Danish, Dutch, English, Estonian, Finnish, French, German, Greek, Hungarian, Irish, Italian, Latvian, Lithuanian, Maltese, Polish, Portuguese, Romanian, Slovak, Slovenian, Spanish, and Swedish) as well as some additional relevant languages (Arabic, Catalan, Chinese, Galician, Hindi, Japanese, Korean, Norwegian, Russian, Turkish, and Ukrainian).

Thus far, we have explored the methodologies for training multilingual LLMs and have developed our initial models: EuroLLM-1.7B and EuroLLM-1.7B-Instruct. This process involved:
\begin{itemize}
    \item Collecting and filtering a large volume of text data for all the targeted languages from various sources, as detailed in \S\ref{sec:data}.
    \vspace{0.2cm}
    \item Defining the mixture of data that composes the training corpus used to train the model. We describe the decisions we took in \S\ref{sec:data}. These decisions were based on scaling laws and on the data availability for each language.
    \vspace{0.1cm}
    \item Developing a multilingual tokenizer, which we depict in \S\ref{sec:tokenizer}.
    \vspace{0.1cm}
    \item Setting the models' hyperparameters and performing pre-training, as described in \S\ref{sec:modeling}.
    \vspace{0.1cm}
    \item Fine-tuning the LLMs to follow natural language instructions, which we describe in \S\ref{sec:post_training}.
    \vspace{0.1cm}
    \item Evaluating the models' performance. Results are reported in \S\ref{sec:results}.
\end{itemize}

\section{Data}
\label{sec:data}
\begin{wrapfigure}[15]{r}{0.5\textwidth}
    \vspace{-0.5cm}
    \includegraphics[width=0.5\textwidth]{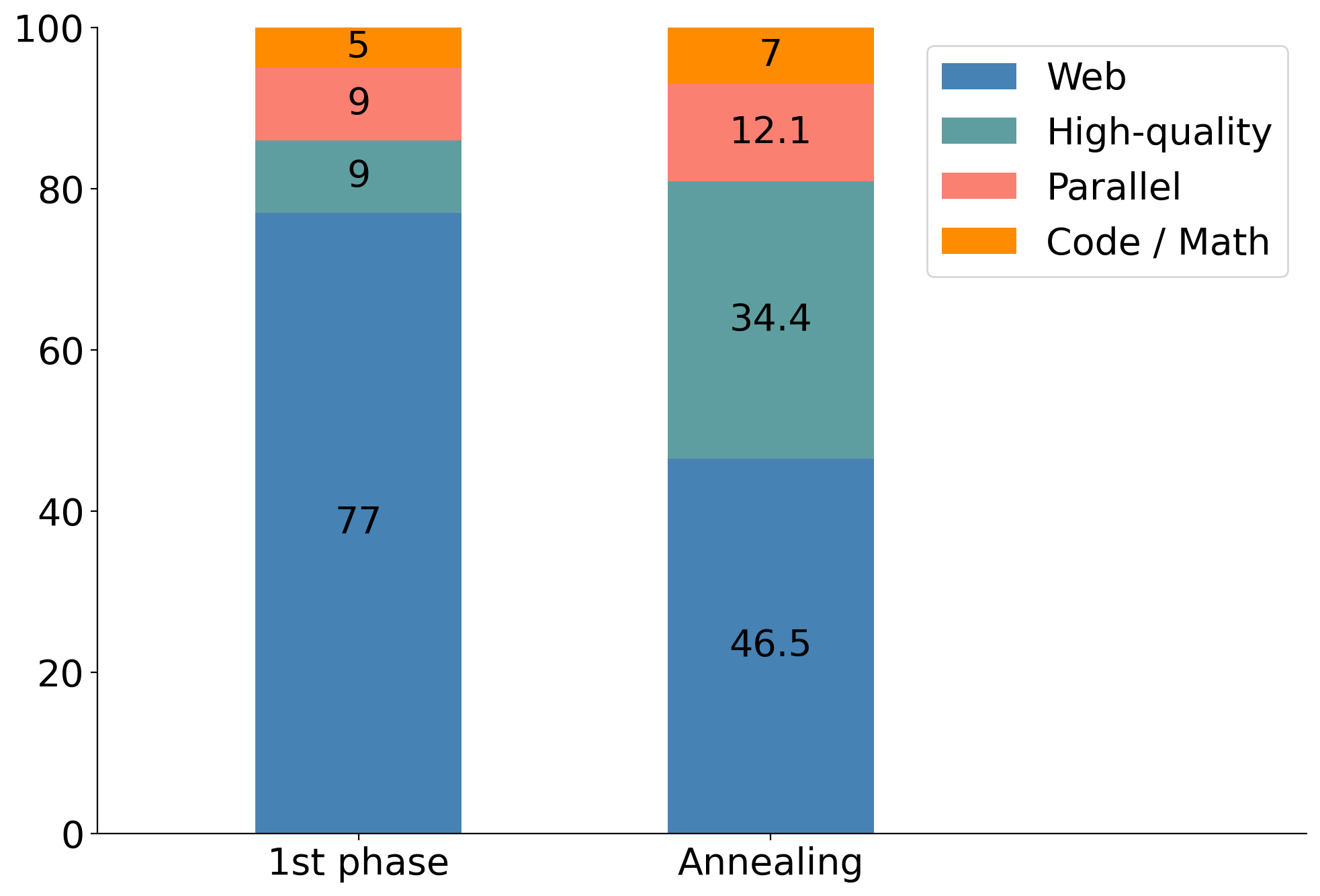}
    \caption{Percentage attributed to each data category in the first training phase (left) and annealing phase (right).}
    \label{fig:percentage_category}
\end{wrapfigure}
To train the EuroLLM models, we collect and filter data from various sources for all supported languages. The data included in the final corpus can be divided into four categories: web data, parallel data, code / math data, and high-quality data. Figure \ref{fig:percentage_category} shows the percentage attributed to each data category. 

\subsection{Data Collection and Filtering}
\paragraph{\textbf{Web Data}}
Regarding web data, for English, we use the FineWeb-edu dataset \citep{lozhkov2024fineweb-edu} which went through individual dump deduplication, heuristic filtering, and model filtering according to the educational level of the documents (we select documents with scores above 2).
For other high-resource languages~(German, Spanish, French, and Italian), we collect data from the RedPajama-Data-v2 \citep{together2023redpajama}, which has been pre-deduplicated. Additionally, we employ a perplexity filter along with a variety of heuristic filters.
For the remaining languages, we collect data from several datasets: HPLT \citep{degibert2024}, MADLAD-400 \citep{kudugunta2023}, CulturaX \citep{nguyen2023}, and mC4 \citep{xue2021}, which we concatenate. We then perform deduplication, language identification, perplexity filtering, and apply a set of heuristic filters, using a preprocessing pipeline based on CCNet \citep{wenzek2019}.

\paragraph{\textbf{Parallel Data}}
Regarding parallel data, we collect to-English (xx→en) and from-English (en→xx) data from various public sources. 
We ensure translation quality by removing sentence pairs below quality thresholds for Bicleaner \citep{prompsit:2018:WMT,prompsit:2020:EAMT} and \textsc{CometKiwi-22} \citep{rei-etal-2022-cometkiwi}.\footnote{For Bicleaner we use a threshold of 0.6 for Portuguese and of 0.5 for all the other languages. For \textsc{CometKiwi-22} we use a threshold of 0.7.}

\paragraph{\textbf{Code / Math Data}}
Regarding code and mathematical data, we collect data from the Stack \citep{Kocetkov2022TheStack}, the Algebraic-stack \citep{azerbayev2023llemma}, and the Open-web-math \citep{paster2023openwebmath} datasets.

\paragraph{\textbf{High-quality Data}}
Regarding high-quality data, we use the Wikipedia \citep{wikidump} for all languages and the Arxiv \citep{clement2019}, Books \citep{Zhu_2015_ICCV}, and Apollo \citep{wang2024apollo} for English.

\paragraph{\textbf{Annealing Data}}
In the last 10\% of the pre-training we increase the predominance of high-quality data in the data mix. To do so, we filter the monolingual data using a binary classifier, inspired by FineWeb-Edu \citep{lozhkov2024fineweb-edu}, which was trained to predict whether a document has some educational value, and collect additional high-quality datasets for this phase: Cosmopedia-v2 \citep{benallal2024smollmcorpus} which is a synthetic dataset composed of textbooks, blog posts, and stories generated by Mixtral-8x7B-Instruct-v0.1 \citep{jiang2024mixtral}; Python-Edu \citep{benallal2024smollmcorpus} which is a subset of Python data from the Stack that was filtered by its educational value; and the training sets of the Grade School Math 8K (GSM8K)  \citep{cobbe2021gsm8k} and of the Mathematics Aptitude Test of Heuristics (MATH) \citep{hendrycksmath2021}.  
We also collect document-level parallel data from Europarl \citep{koehn2005} and ParaDocs \citep{wicks2024}.

\subsection{Data Mixture}
Before starting the training of multilingual LLMs, it is crucial to carefully define the data mixture to be used. This involves deciding how much parallel data to include (\S\ref{sec:parallel_data}), determining whether to repeat high-quality data (\S\ref{sec:high_quality_data}), and deciding how to allocate the total number of tokens among the different languages (\S\ref{sec:language_division}).

\subsubsection{Parallel Data}
\label{sec:parallel_data}
Parallel data (sentences / documents with their translations in another language) can benefit multilingual LLMs in two aspects: improving the alignment between languages and enhancing the model's machine translation capabilities. However, determining the optimal proportion of parallel data can be challenging.

\paragraph{\textbf{Joint Scaling Laws}} Recent research suggests that the performance of large LLMs can be predicted by a function of the number of non-embedding parameters $N$, using a power-law \citep{kaplan2020scaling}.
In particular, \cite{fernandes2023scaling} found that, for \textit{multilingual} models, by training smaller models with varying weights for each language in the data mix, one can fit a \textit{multilingual, joint} scaling law that predicts performance for a model trained with $p$ weight for a language:
$\mathcal{L}(N, p) = f(p) \beta  N^{-\alpha} + L_\infty$, with a \textit{ratio} function $f(p) = p + c_1 p^{c_2}(1 - p)^{c_3}$, and where $\alpha$, $\beta$, $L_\infty$ and $c_{\{1,2,3\}}$ are empirically estimated parameters of the scaling law. 
This law can predict the language performance trade-off of larger models, even for novel language weightings not encountered during the fitting of the scaling law. 

\begin{figure}[t]
    \centering
    \includegraphics[width=\textwidth]{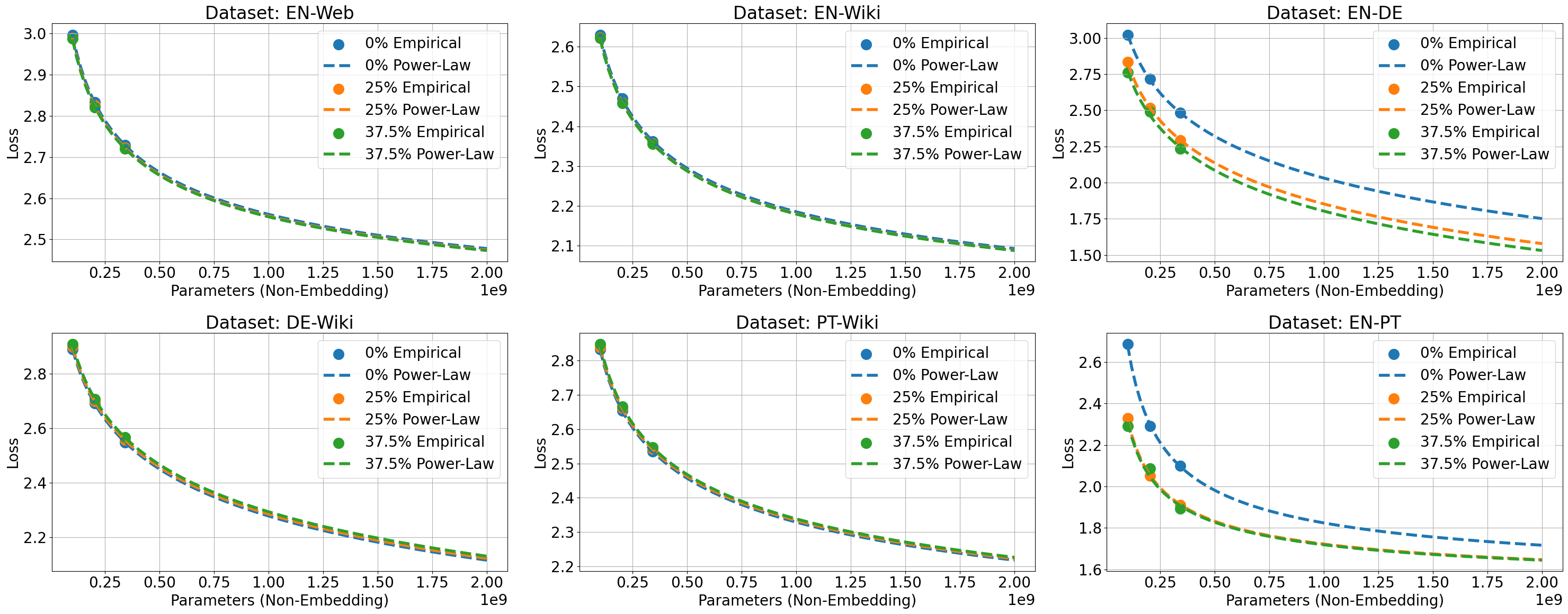}
    \caption{Joint Scaling laws obtained when varying the percentage of parallel data.}
    \label{fig:scaling_law_parallel}
\end{figure}

Thus, to decide on the appropriate amount of parallel data, we re-purpose this law to predict the impact on performance as we change its \textit{weighting} in training: we train models with varying numbers of non-embedding parameters (100M, 203M, and 341M) on a 100B token corpus, for which parallel data constitutes different percentages (0\%, 25\%, and 37.5\%) of the total data for each language, excluding English.
Figure \ref{fig:scaling_law_parallel} reports the obtained scaling laws for test sets from several domains: web data, Wikipedia data, and parallel data. 
The results indicate that adding parallel data does not negatively impact performance on web and Wikipedia domains, while significantly enhancing the performance on parallel data. 
Moreover, increasing the percentage of parallel data from 25\% to 37.5\% yields diminishing returns. Therefore, we include 20\% parallel data for each language in the final corpus.

\subsubsection{Repeating High Quality Data}
\label{sec:high_quality_data}
To determine whether it is beneficial to repeat datasets considered to be of higher quality, we analyze scaling laws using a method similar to that described in \S\ref{sec:parallel_data}. To do so, we train models on two 100B token corpora: one where Wikipedia data is repeated for all languages and one where it is not.

\begin{figure}[t]
    \centering
    \includegraphics[width=\textwidth]{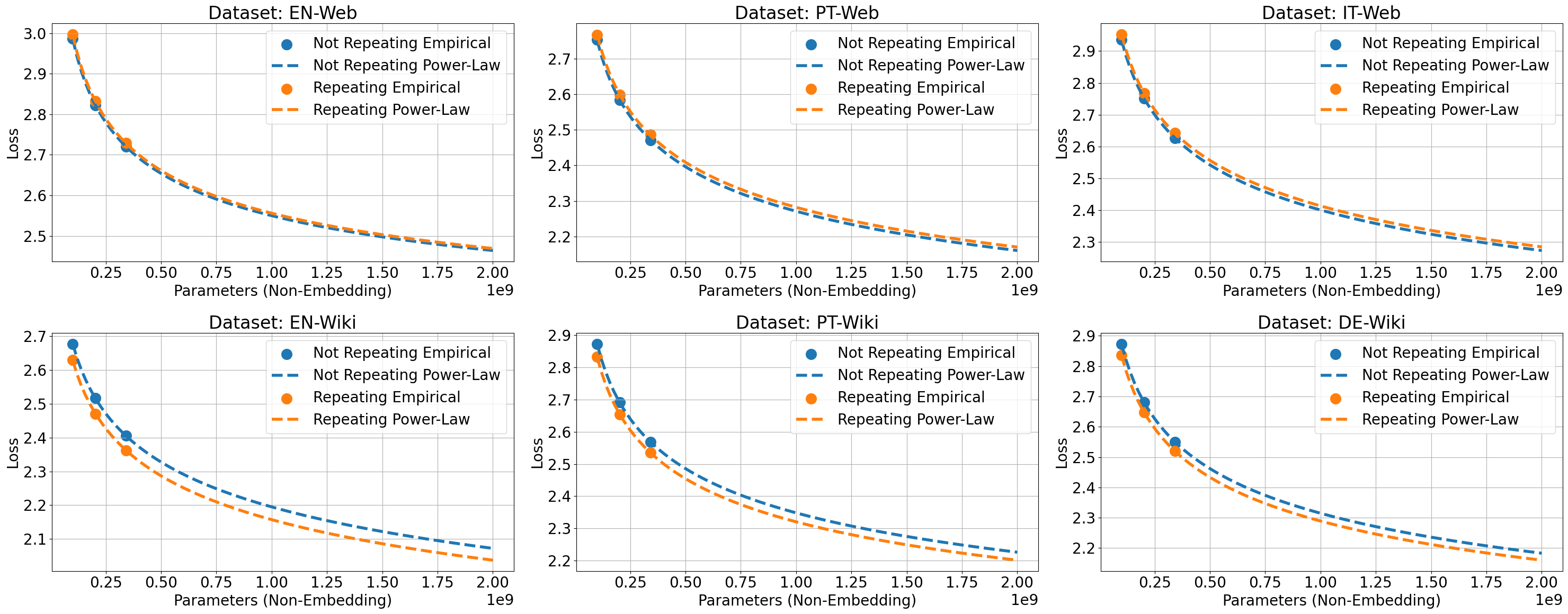}
    \caption{Joint Scaling laws obtained when repeating vs not-repeating Wikipedia.}
    \label{fig:scaling_law_wiki}
\end{figure}

Figure \ref{fig:scaling_law_wiki} shows the scaling laws for test sets from web and Wikipedia domains. The results clearly indicate that repeating Wikipedia data improves performance on the Wikipedia test sets without degrading performance on the web test sets. Therefore, we choose to repeat data from high-quality datasets.

\subsubsection{Division between Languages}
\label{sec:language_division}
\begin{figure}[t]
    \centering
    \includegraphics[width=\textwidth]{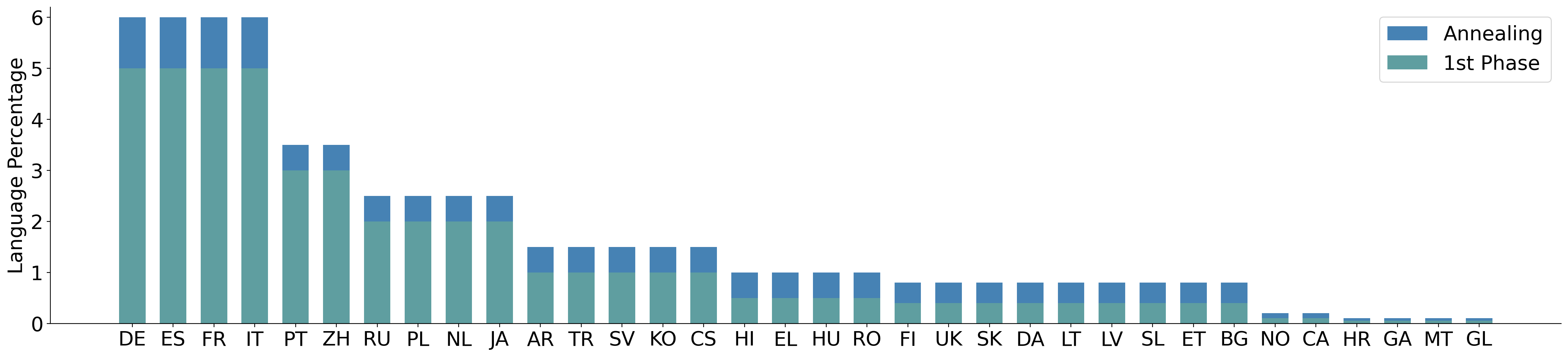}
    \caption{Percentage of the training corpus attributed to each language, excluding English which accounts to 50\% in the first phase and 32.5\% during annealing. 5\% of the corpus is left for datasets composed of code and math in the first phase and 7\% during annealing.}
    \label{fig:language_percentage}
\end{figure}
Regarding the allocation of the corpus to each language, we designate 50\% for English, as both high-quality data and web data are predominantly in English, and include 5\% of code / math data. The remaining 45\% of the tokens are distributed among the other languages based on the amount of data obtained after the collection and filtering processes. 
In order to increase EuroLLM's multilinguality, in the annealing phase, we decrease the English allocation to 32.5\% and distribute the surplus across the other languages. We also increase the code / math allocation to 7\%. 
Figure \ref{fig:language_percentage} shows the exact percentage attributed to each language.

\section{Tokenizer}
\label{sec:tokenizer}
To train the tokenizer, we adopt the approach used by the LLaMa-2 and Mistral models~\citep{touvron2023llama,jiang2023mistral}, training a BPE tokenizer with byte-fallback. To do so, we use the SentencePiece framework~\citep{kudo2018sentencepiece}.
For an LLM to be efficient across a large number of languages, it is crucial to have a tokenizer with a large vocabulary. However, this comes with the drawback of having a high number of embedding parameters. Through experimentation, we reach the conclusion that a vocabulary of 128,000 pieces provides the best trade-off.

\begin{figure}[h]
    \centering
    \includegraphics[width=\textwidth]{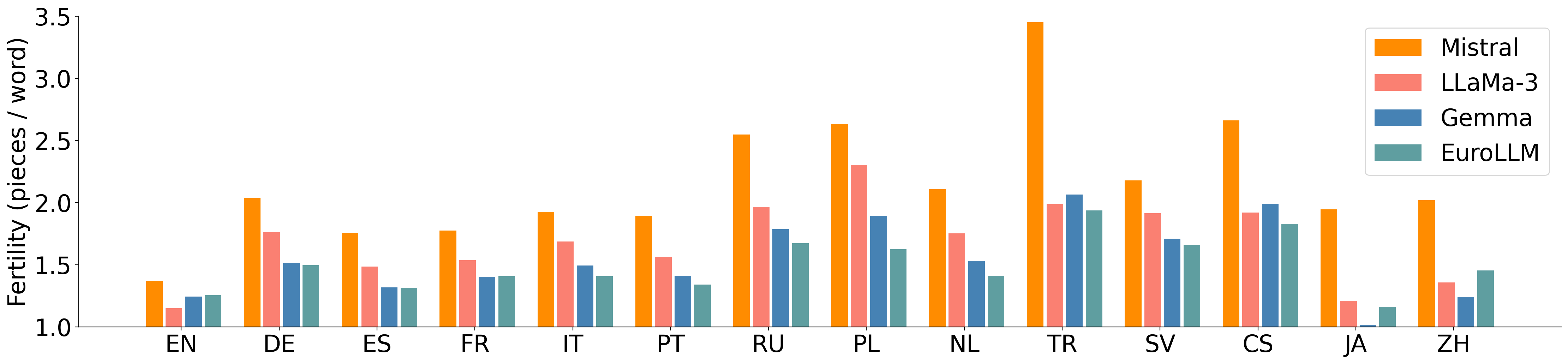}
    \caption{Fertility (pieces / word) obtained with the Mistral, LLaMa-3, Gemma, and EuroLLM tokenizers for a subset of the EuroLLM languages.}
    \label{fig:fertility_comparison}
\end{figure}

We compare the fertility achieved by the EuroLLM tokenizer with those of Mistral, LLaMa-3, and Gemma tokenizers \citep{jiang2023mistral,llama3modelcard,team2024gemma} which have vocabularies of 32,000, 128,256, and 256,000 pieces, respectively. Figure \ref{fig:fertility_comparison} presents the fertilities for a subset of the languages included in EuroLLM. Compared to the Mistral tokenizer, the larger vocabulary of EuroLLM results in significantly lower fertilities. In comparison with the LLaMa-3 and Gemma tokenizers, the LLaMa-3 tokenizer shows the lowest fertility in English but higher fertility for most other languages, while the Gemma tokenizer seems to be better for Asian languages but very similar to EuroLLM for the European ones.

\section{Modeling}
\label{sec:modeling}
\begin{wraptable}[20]{r}{0.5\textwidth}
\small
\vspace{-0.4cm}
\setlength{\tabcolsep}{1.5ex}
\begin{tabular}{lc}
\toprule
         & \textbf{1.7B} \\ \midrule
Sequence Length & 4,096 \\
Number of Layers & 24 \\
Embedding Size & 2,048 \\
FFN Hidden Size & 5,632 \\
Number of Heads & 16 \\
Number of KV Heads (GQA) & 8 \\
Activation Function & SwiGLU \\
Position Encodings & RoPE ($\Theta$=10,000) \\
Layer Norm & RMSNorm \\
Tied Embeddings & No \\
Max Learning Rate & $3\times 10^{-4}$ \\
Min Learning Rate & $3\times 10^{-5}$ \\
\midrule
Embedding Parameters & 0.262B \\
LM Head Parameters & 0.262B \\
Non-embedding Parameters & 1.133B \\
Total Parameters & 1.657B \\
\bottomrule
\end{tabular} 
\caption{Overview of EuroLLM hyperparameters.}
\label{tab:hyperparameters}
\end{wraptable}
EuroLLM uses a standard, dense Transformer architecture \citep{vaswani2017attention}:
\begin{itemize}[leftmargin=0.5cm, itemsep=0pt]
    \item We use grouped query attention (GQA; \cite{ainslie2023gqa}) with 8 key-value heads since it has been shown to increase speed at inference time while maintaining downstream performance \citep{team2024gemma2}.
    \vspace{0.1cm}
    \item We use pre-layer normalization \citep{xiong2020layer}, since it improves training stability, and use the RMSNorm \citep{zhang2019root}, which is faster than LayerNorm \citep{ba2016layer}.
    \vspace{0.1cm}
    \item We use the SwiGLU activation function \citep{shazeer2020glu} since it has been shown to lead to good results on downstream tasks \citep{shazeer2020glu,le2022language}.
    \vspace{0.1cm}
    \item We use rotary positional embeddings (RoPE) \citep{su2024roformer} in every layer since these have been shown to lead to good performances while allowing the extension of the context length. 
\end{itemize}

\paragraph{\textbf{Training}} We pre-train EuroLLM-1.7B on 4 trillion tokens, increasing the predominance of high-quality data on the final 10\% of the pre-training process.
We use 256 Nvidia H100 GPUs of the Marenostrum 5 supercomputer, training the model with a constant batch size of 3,072 sequences, which corresponds to approximately 12 million tokens, using the Adam optimizer \citep{kingma2014adam}, and \texttt{bfloat16} mixed precision. All relevant model and training hyperparameters are shown in Table~\ref{tab:hyperparameters}.

\subsection{Learning Rate Scheduler}
Regarding the learning rate scheduler, we experiment with two options. In the first option, we use a cosine scheduler with a warm-up phase corresponding to 10\% of the steps. The second option consists of using a trapezoid scheduler \citep{xing2018walk} (also named Warmup-Stable-Decay \citep{hu2024minicpm}). This scheduler has three phases: warm-up for 10\% of the steps; constant learning rate; linear decay of the learning rate to the minimum learning rate in the final 10\% of the pre-training process, a phase in which we use the higher quality annealing data. To decide which option to use in future models we compare the two options on two multilingual general benchmarks: Hellaswag \citep{zellers2019hellaswag,lai2023okapi} and Arc Challenge \citep{allenai:arc,lai2023okapi} and on machine translation on the \textsc{Flores-200} \citep{nllb2022}, WMT-23 \citep{kocmi2023findings}, and WMT-24 \citep{kocmi2024preliminary} datasets. The machine translation scores are obtained using \textsc{Comet-22} \citep{rei2022comet}.
The average results, reported on Table \ref{tab:cosine_trapezoid}, show that using the trapezoid scheduler leads to scores consistently better on the multilingual benchmarks and on machine translation.
\begin{table}[h]
\small
\setlength{\tabcolsep}{1.8ex}
\begin{tabular}{lccccc}
\toprule
 & \multicolumn{2}{c}{\textbf{\textsc{General}}} & \multicolumn{3}{c}{\textbf{\textsc{Translation}}}\\
\textsc{Model} & Hellaswag & Arc Challenge & \textsc{Flores-200} & WMT-23 & WMT-24 \\ \midrule
EuroLLM-1.7B - cosine & 0.4646 & 0.3206 & 86.48 & 82.88 & 78.87 \\
EuroLLM-1.7B - trapezoid & \textbf{0.4744} & \textbf{0.3268} & \textbf{86.75} & \textbf{83.13} & \textbf{79.35} \\
\bottomrule
\end{tabular} 
\caption{Comparison between models trained using the two learning rate scheduler options: cosine scheduler and trapezoid scheduler.}
\label{tab:cosine_trapezoid}
\end{table}

\section{Post Training}
\label{sec:post_training}

\paragraph{\textbf{EuroBlocks}} In order for EuroLLM-1.7B to be able to follow natural language instructions, we create a multilingual dataset --- EuroBlocks --- which encompasses publicly available human-written and synthetic data. We use instruction-following conversations collected from OpenHermes-2.5 \citep{OpenHermes_2.5} and Aya \citep{singh2024aya} datasets, as well as high-quality machine translation examples from NTREX-128 \citep{federmann-etal-2022-ntrex}, \textsc{Flores-200-dev} \citep{nllb2022}, WMT-21 \citep{farhad2021findings}, and WMT-22 \citep{kocmi2022findings}. Overall the dataset is composed by 1M samples covering all supported languages and a variety of tasks. 

\paragraph{\textbf{Supervised fine-tuning (SFT)}} We fine-tune EuroLLM-1.7B on EuroBlocks to turn it into an instruction-following conversational model: EuroLLM-1.7B-Instruct. We use the standard cross-entropy loss, enabling \texttt{bfloat16} mixed precision and packing. We only calculate the loss on target tokens~(thus masking loss on prompt tokens). We train for 4 epochs using a learning rate of $7 \cdot 10^{-6}$ over the course of around 6K steps. 

\paragraph{\textbf{Formatting}} EuroLLM is finetuned with chat format control tokens~(\textcolor{blue}{\texttt{<|im\_start|>}} and \textcolor{blue}{\texttt{<|im\_end|>}} tokens) that are included in the original tokenizer described in Section~\ref{sec:tokenizer}. As such, no tokens are added during SFT; however, we overwrite the end-of-sequence token with the \textcolor{blue}{\texttt{<|im\_end|>}} token. These control tokens create a clear separation between user/model turns. The chat template may also include a "system prompt" that can be used for added functionality and steerability of the model's responses. We show an example in Table~\ref{tab:dialogue_example}.

\begin{table}[t]
\begin{center}  
\footnotesize
\begin{tabular}{p{1cm}p{12cm}}
\toprule
\textbf{System}   & \texttt{<s>}{\textcolor{blue}{\texttt{<|im\_start|>system}}} \\\cdashlinelr{1-2}
                 & Translate all user texts to English.{\textcolor{blue}{\texttt{<|im\_end|>}}} \\
\textbf{User}   & {\textcolor{blue}{\texttt{<|im\_start|>user}}} \\
                 & A Comissão Europeia é uma instituição politicamente independente e que representa e defende os interesses da União Europeia na sua globalidade. {\textcolor{blue}{\texttt{<|im\_end|>}}} \\
                 & {\textcolor{blue}{\texttt{<|im\_start|>assistant}}} \\
\textbf{Model}   & The European Commission is a politically independent institution that represents and defends the interests of the European Union as a whole.
{\textcolor{blue}{\texttt{<|im\_end|>}}}\\\cdashlinelr{1-2}
\textbf{User}   & {\textcolor{blue}{\texttt{<|im\_start|>user}}} \\
                 & La Comisión Europea no consta únicamente de los 27 miembros que integran la institución constitucional, el llamado Colegio de Comisarios. Está integrada por todo un cuerpo administrativo de algo más de 38 000 funcionarios cuya función es asistirlo y actuar como brazo ejecutivo del Colegio.
 {\textcolor{blue}{\texttt{<|im\_end|>}}} \\
                 & {\textcolor{blue}{\texttt{<|im\_start|>assistant}}} \\
\textbf{Model}   & The European Commission is not only composed of the 27 members who form the constitutional institution, the so-called College of Commissioners. It is composed of a whole administrative body of more than 38,000 officials whose function is to assist it and act as the executive arm of the College.
{\textcolor{blue}{\texttt{<|im\_end|>}}}\\
\bottomrule
\end{tabular}
\end{center}
\caption{Example of a dialogue with EuroLLM-1.7B-Instruct. We mark the system, user and model control tokens.}
\label{tab:dialogue_example}
\end{table}

\section{Results}
\label{sec:results}
In order to evaluate our models we recur to: general benchmarks (\S\ref{sec:general_bench}) to understand if EuroLLM-1.7B acquired world knowledge during training; and machine translation (\S\ref{sec:mt}) to understand if the model is able to understand and generate text in all the supported languages.

\subsection{General Benchmarks}
\label{sec:general_bench}
\begin{figure}[t]
    \centering
    \includegraphics[width=\textwidth]{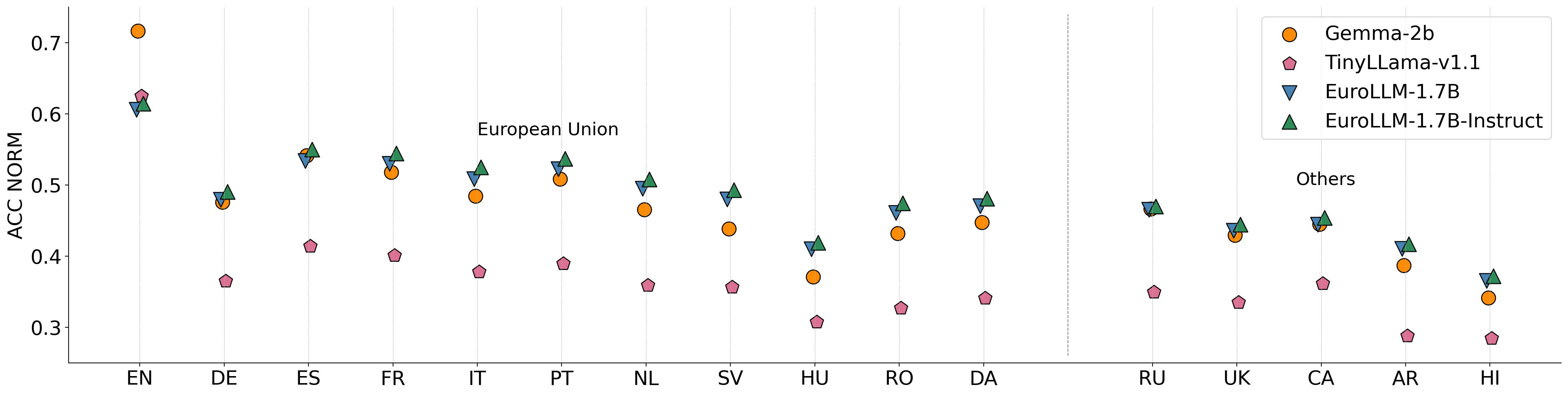}
    \vspace{-0.3cm}

    \includegraphics[width=\textwidth]{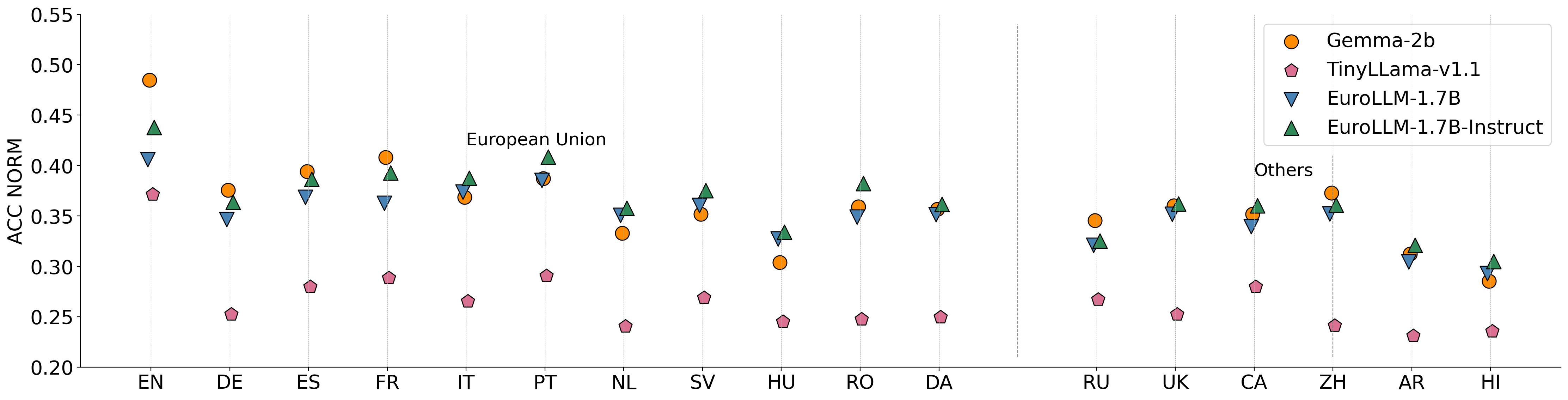}
    \caption{Results on the Hellaswag (top) and Arc Challenge (bottom) benchmarks. The results were obtained using 10-shot and 25-shot prompts for Hellaswag and Arc Challenge, respectively. }
    \label{fig:results_heallaswag}
\end{figure}
Regarding general benchmarks, we consider a commonsense natural language inference test set: Hellaswag \citep{zellers2019hellaswag} and a test set of science exam questions: Arc Challenge \citep{allenai:arc}. These benchmarks are originally English-only, so we recur to translations \citep{lai2023okapi}. 
As baselines, we use Gemma-2b \citep{team2024gemma} and TinyLlama \citep{zhang2024tinyllama}. 
Figure \ref{fig:results_heallaswag} reports the results. On Hellaswag, EuroLLM-1.7B matches or outperforms Gemma-2b and TinyLlama on all languages besides English, which showcases its increased multilinguality.
On Arc Challenge, EuroLLM-1.7B outperforms TinyLlama on all languages but is worse than Gemma-2b. This can be caused by the lower number of parameters (EuroLLM-1.7B has 1.133B non-embedding parameters while Gemma-2B has 1.981B).
Interestingly, EuroLLM-1.7B-Instruct leads to slightly better results than EuroLLM-1.7B for both benchmarks.

\subsection{Machine Translation}
\label{sec:mt}
\begin{figure}[t]
    \centering
    \includegraphics[width=\textwidth]{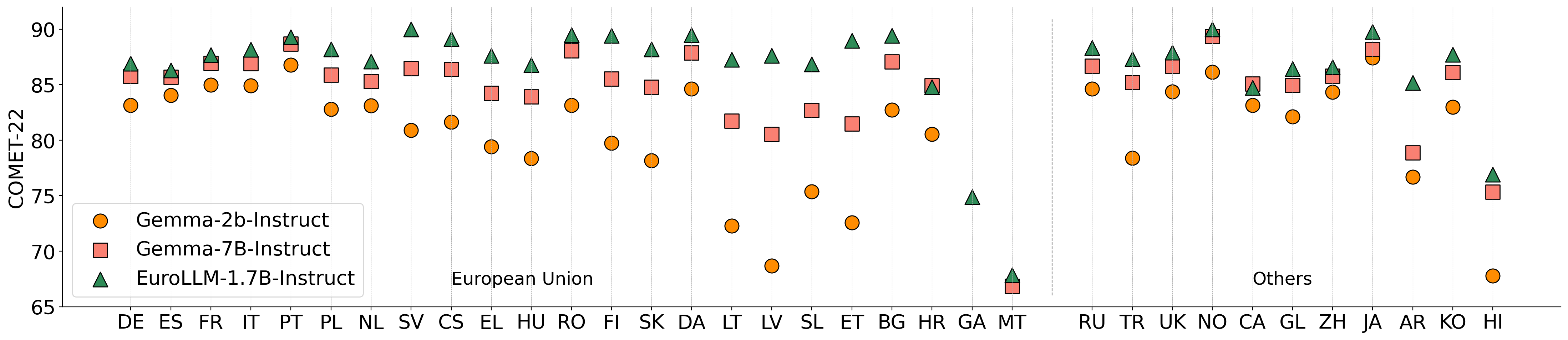}
    \vspace{-0.3cm}
    
    \includegraphics[width=\textwidth]{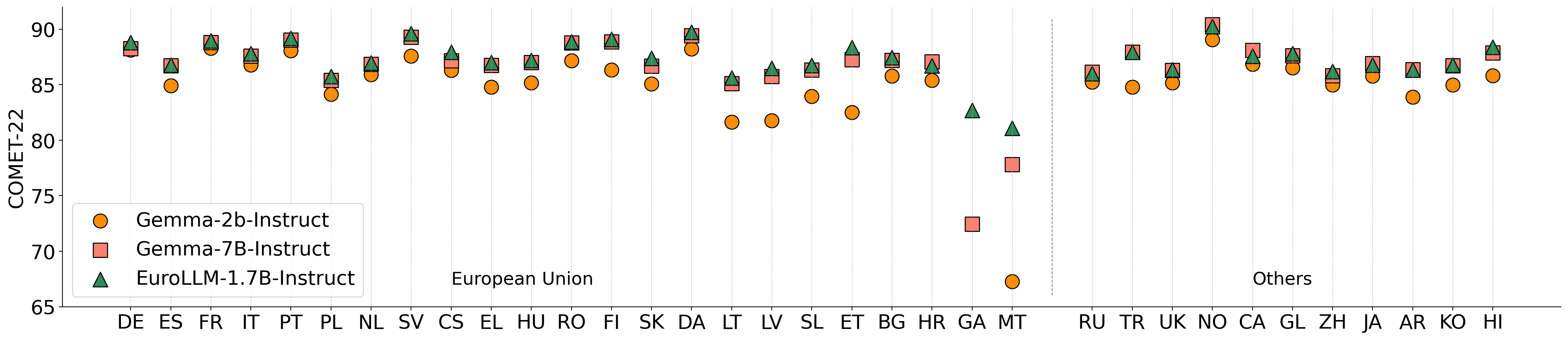}
    \caption{\textsc{Comet-22} scores on the \textsc{Flores-200} dataset on EN-XX (top) and XX-EN (bottom) language pairs. All models were fine-tuned with the EuroBlocks dataset and the translations were obtained using 0-shot prompts and greedy search.}
    \label{fig:results_flores}
\end{figure}
\begin{figure}[t]
    \centering
    \includegraphics[width=\textwidth]{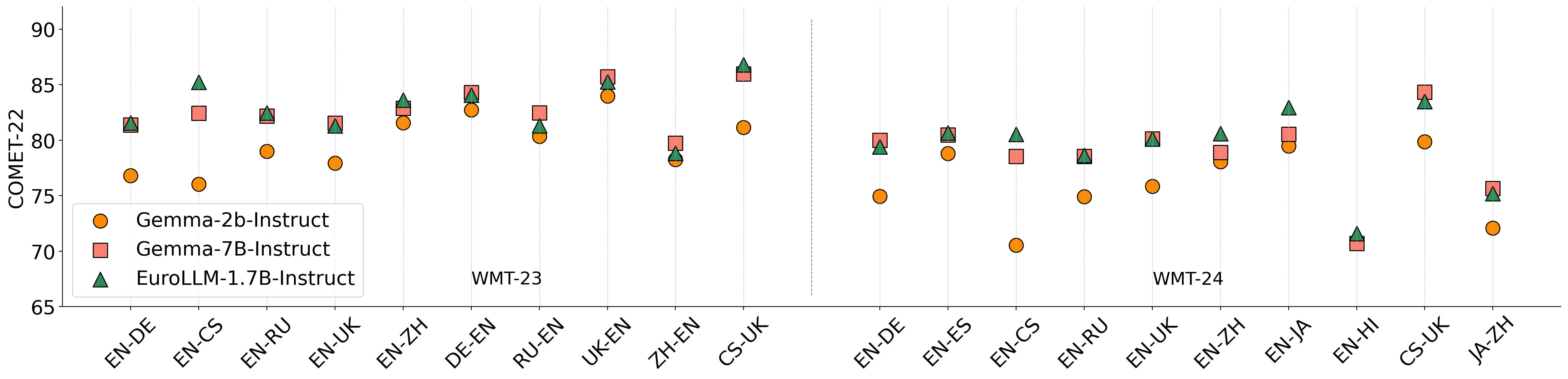}
    \caption{\textsc{Comet-22} scores on the WMT-23 and WMT-24 datasets. All models were fine-tuned with the EuroBlocks dataset and the translations were obtained using 0-shot prompts and greedy search.}
    \label{fig:results_wmt}
\end{figure}

Regarding machine translation, we compare EuroLLM-1.7B-Instruct with Gemma-2b and Gemma-7b \citep{team2024gemma} on three datasets: \textsc{Flores-200-test} \citep{nllb2022}, WMT-23 \citep{kocmi2023findings}, and WMT-24 \citep{kocmi2024preliminary} and evaluate the translations using \textsc{Comet-22}. To have a fair comparison, we also fine-tune Gemma-2b and Gemma-7b on the EuroBlocks dataset. 
Figures \ref{fig:results_flores} and \ref{fig:results_wmt} report the results. We can see that EuroLLM-1.7B-Instruct clearly outperforms Gemma-2b-Instruct on all languages pairs and datasets, and is competitive with Gemma-7b-Instruct despite the much lower number of parameters.

\section{Conclusions and Future Work}
In this paper, we present the work done so far in the EuroLLM project. We describe our data collection and filtering process, how we build a multilingual tokenizer, and the data mixture and modeling configurations. We also release our initial models: EuroLLM-1.7B and EuroLLM-1.7B-Instruct and report their performance on multilingual general benchmarks and machine translation. 
In future work, we will continue training multilingual LLMs for Europe, focusing on scaling up the number of model parameters and improving further the quality of our data.

\subsubsection*{Acknowledgments}
Part of this work was supported by the EU’s Horizon Europe Research and Innovation Actions (UTTER, contract 101070631), by the project DECOLLAGE (ERC-2022-CoG
101088763), and by the Portuguese Recovery and Resilience Plan through project C645008882-00000055 (Center for Responsible AI). We thank EuroHPC for the HPC resources used to support this work through grant EHPC-EXT-2023E01-04.

\newpage
\bibliography{iclr2025_conference}
\bibliographystyle{iclr2025_conference}

\end{document}